\documentclass[runningheads]{llncs}
\usepackage[T1]{fontenc}
\usepackage{graphicx}
\usepackage{booktabs}
\usepackage[misc]{ifsym}
\newcommand{\corr}{(\Letter)}
\usepackage{float}
\usepackage{subcaption,xcolor,lipsum}
\usepackage{tabularx}
\usepackage{multicol}
\usepackage{multirow}
\newcolumntype{L}[1]{>{\hsize=#1\hsize\raggedright\arraybackslash}X}
\newcolumntype{C}[1]{>{\hsize=#1\hsize\centering\arraybackslash}X}
\newcolumntype{R}[1]{>{\hsize=#1\hsize\raggedleft\arraybackslash}X}
\usepackage{array}
\usepackage{tabularray} 
\newcolumntype{M}[1]{>{\centering\arraybackslash}m{#1}} 
\newcolumntype{P}[1]{>{\raggedright\arraybackslash}p{#1}} 
\newcolumntype{B}[1]{>{\raggedright\arraybackslash}b{#1}} 

\usepackage{pifont}
\newcommand{\cmark}{\ding{51}}%
\usepackage{amsmath,amssymb}
\usepackage{wasysym} 
\usepackage{enumerate}
\usepackage{hyperref}
\usepackage[depth=2]{bookmark}

\begin{document}

\title{Task-Agnostic Contrastive Pretraining for\\Relational Deep Learning }

\titlerunning{Task-Agnostic Contrastive Pretraining for Relational Deep Learning}

\author{
Jakub Pele\v{s}ka  \corr \and
Gustav \v{S}\'{\i}r 
}

\authorrunning{J. Pele\v{s}ka and G. \v{S}\'{\i}r}

\institute{
Czech Technical University in Prague,\\
Karlovo náměstí 13, Prague, 121 35, Czechia
\email{jakub.peleska@fel.cvut.cz,gustav.sir@cvut.cz}
}


\maketitle              

\begin{abstract}

Relational Deep Learning (RDL) is an emerging paradigm that leverages Graph Neural Network principles to learn directly from relational databases by representing them as heterogeneous graphs. However, existing RDL models typically rely on task-specific supervised learning, requiring training separate models for each predictive task, which may hamper scalability and reuse. 

In this work, we propose a novel task-agnostic contrastive pretraining approach for RDL that enables database-wide representation learning. For that aim, we introduce three levels of contrastive objectives---row-level, link-level, and context-level---designed to capture the structural and semantic heterogeneity inherent to relational data. We implement the respective pretraining approach through a modular RDL architecture and an efficient sampling strategy tailored to the heterogeneous database setting. Our preliminary results on standard RDL benchmarks demonstrate that fine-tuning the pretrained models measurably outperforms training from scratch, validating the promise of the proposed methodology in learning transferable representations for relational data.



\end{abstract}

\section{Introduction}

From their establishment~\cite{Codd1970}, \textit{Relational Databases} (RDBs) have played a pivotal role in ushering our society into the information age. By storing data as interconnected tables safeguarded by integrity constraints, RDBs provide a robust and highly expressive framework for managing structured information. As a result, they remain a cornerstone of critical systems across a broad spectrum of domains, ranging from healthcare~\cite{white_pubmed_2020} to government~\cite{maali_enabling_2010}.

Despite their widespread adoption, RDBs are inherently misaligned with conventional Machine Learning (ML) pipelines, which assume data in the standard form of fixed-size, independent and identically distributed (i.i.d.) feature vectors—commonly referred to as the ``tabular'' learning format. In contrast, RDBs contain multiple interrelated tables of varying sizes, violating this assumption. To bridge the gap, traditional approaches have typically relied on \textit{propositionalization}''~\cite{propos}, which is essentially a feature extraction process that aggregates relational substructures into flattened attributes (features) of the tabular format. However, this transformation inevitably results in a loss of structural and semantic information that is often crucial for relational learning.

Recent advances in \textit{Graph Representation Learning}~\cite{hamilton_graph_2020} have enabled an alternative approach. By representing an RDB as a heterogeneous (and potentially temporal) graph, where each row becomes a node and inter-table relationships are captured by edges derived from integrity constraints, it becomes possible to apply the ``message-passing'' principles of \textit{Graph Neural Network} (GNN)\cite{wu2020comprehensive} to relational data. This line of work has led to the recent emergence of the \textit{Relational Deep Learning} (RDL)~\cite{fey2024position} field, which adapts various GNN-based message-passing schemes to the relational setting, achieving some promising initial results across a variety of supervised tasks~\cite{Cvitkovic2020,zhang2023gfs,zahradnik2023deep,peleska_transformers_2024}.

However, most existing RDL methods are designed around specific downstream tasks and require training separate models for each, limiting their scalability and reusability in practical, multi-purpose database applications. To address this gap, we propose a novel contrastive pretraining methodology for RDL that enables general-purpose, task-agnostic representation learning over relational databases. At the core of the approach, we introduce a three-level contrastive objective---operating at the row, link, and context levels of the database graph---that captures both attribute semantics and structural dependencies. The resulting pretrained models can then be effectively fine-tuned for diverse downstream tasks, reducing the need for repeated task-specific training. We implement this approach within a custom modular RDL architecture and demonstrate its effectiveness on standard relational benchmarks.
Our initial experimental results demonstrate that fine-tuning the pretrained models consistently and measurably outperforms training from scratch, validating the effectiveness of our approach in learning transferable representations for relational data, opening doors for exploring a new frontier of foundational database model training.



\section{Background}

This paper builds on learning from RDBs (Sec.~\ref{sec:rdb}) with GNN-based models (Sec.~\ref{sec:gnn}), forming the backbone of the RDL paradigm (Sec.~\ref{sec:rdl}).


\subsection{Relational Databases}
\label{sec:rdb}

Principles of RDBs are formally based on the \textit{relational model}~\cite{codd1990relational}, which is grounded in relational logic~\cite{gallier2015logic}. This abstraction enables the definition of any database, regardless of specific software implementation, as a collection of $n$-ary relations, which are defined over the domains of their respective attributes, managed by the Relational Database Management System (RDBMS) to ensure data consistency with the integrity constraints of the database schema.
The key concepts to be used in this paper are as follows.

\subsubsection{Relational Database} A Relational Database (RDB) $\mathcal{R}$ is defined as a finite set of relations $R_1, R_2, \dots ,R_n$. An instance of an RDB $\mathcal{R}$ is implemented through a RDBMS, enabling to perform Structured Query Language (SQL;~\cite{chamberlin_sequel_1974}) operations, rooted in {relational algebra}.

\subsubsection{Relation (Table)} Formally, an $n$-ary relation $R_{/n}$ is a subset of the Cartesian product defined over the domains $D_i$ of its $n$ \textit{attributes} $A_i$ as $R_{/n} \subseteq D_1 \times D_2 \times \dots \times D_n$, where $D_i = \mathsf{dom}(A_i)$. Each relation $R$ consists of a heading (signature) $R_{/n}$, formed by the set of its attributes, and a body, formed by the values of the respective attributes, commonly represented as a \textit{table} $T_R$ of the relation $R$.

\subsubsection{Attribute (Column)} \textit{Attributes} $\mathcal{A}_R = \{A_1, \ldots, A_n\}$ define the terms of a relation $R_{/n}$, corresponding to the \textit{columns} of the respective table $T_R$. Each attribute is a pair of the attribute's name and a \textit{type}, constraining the domain of each attribute as $\mathsf{dom}(A_i) \subseteq \mathsf{type}(D_i)$. An attribute \textit{value} $a_i$ is then a specific valid value from the respective domain of the attribute $A_i$.

\subsubsection{Tuple (Row)} An $n-$\textit{tuple} in a relation $R_{/n}$ is a tuple
of attribute values ${t_i} = (a_1, a_2, \ldots, a_n)$, where $a_j$ represents the value of the attribute $A_j$ in $R$. The relation can thus be defined extensionally by the \textit{unordered} set of its tuples: $R = \{t_1, t_2,\ldots, t_m\}$, corresponding to the \textit{rows} of the table $T_R$.

\subsubsection{Integrity constraints} In addition to the domain constraints $\mathsf{dom}(A_i)$, the most important integrity constraints are the primary and foreign keys. A \textit{primary} key $PK$ of a relation $R$ is a minimal subset of its attributes $R[PK] \subseteq \mathcal{A_R}$ that uniquely identifies each tuple: 
$\forall t_1, t_2 \in R:~ (t_1[PK] = t_2[PK]) \Rightarrow (t_1 = t_2)$. 
A~\textit{foreign} key ${FK}_{R_2}$ in relation $R_1$ then refers to the primary key ${PK}$ of another relation $R_2$ as 
$\forall t \in R_1:~ t[FK] \in \{t'[PK] \mid t' \in R_2\} \,.$
This constitutes the inter-relations in the database, with the RDBMS handling the \textit{referential integrity} of ${T_{R_1}}[FK] \subseteq {T_{R_2}}[PK]$.


\subsection{Graph Neural Networks}
\label{sec:gnn}
Graph Neural Networks constitute a comprehensive class of neural models designed to process graph-structured data through the concept of (differentiable) \textit{message-passing}~\cite{wu2020comprehensive}. Given an input graph $G = (\mathcal{V}, \mathcal{E})$, with a set of nodes $\mathcal{V}$ and edges $\mathcal{E}$, let $h_v^{(l)} \in \mathbb{R}^{d^{(l)}}$ be the vector representation (embedding) of node $v$ at layer $l$.
The general concept of GNNs can then be defined through the following sequence of three functions:

\begin{enumerate}
    \item[(i)] \textit{Message} function $M^{(l)}: \mathbb{R}^{d^{(l)}} \times \mathbb{R}^{d^{(l)}} \to \mathbb{R}^{d_m^{(l)}}$ computes messages for each edge $(u, v) \in E$ as 
        $m_{u \to v}^{(l)} = M^{(l)}(h_u^{(l)}, h_v^{(l)}) \,.$
    
    \item[(ii)] \textit{Aggregation} function $A^{(l)}: \{\mathbb{R}^{d_m^{(l)}}\} \to \mathbb{R}^{d_m^{(l)}}$ aggregates the messages for each $v \in V$ as 
       $M_v^{(l)} = A^{(l)}\left(\{m_{u \to v}^{(l)} ~|~ (u, v) \in E\}\right) \,.$
    
    \item[(iii)] \textit{Update} function $U^{(l)}: \mathbb{R}^{d^{(l)}} \times \mathbb{R}^{d_m^{(l)}} \to \mathbb{R}^{d^{(l+1)}}$ updates representation of each $v \in V$ as 
        $h_v^{(l+1)} = U^{(l)}(h_v^{(l)}, M_v^{(l)}) \,.$
\end{enumerate}

The specific choice of message, aggregation, and update functions varies across specific GNN models, which are typically structured with a predefined number $L$ of such layers, enabling the message-passing to propagate information across $L$-neighborhoods within the graph(s).

\subsection{Relational Deep Learning}
\label{sec:rdl}
In this paper, we adopt the concept of RDL as extending mainstream deep learning models, particularly the GNNs (Sec.~\ref{sec:gnn}), for application to RDBs (Sec.~\ref{sec:rdb}). For completeness, in the relational learning community~\cite{cropper2020turning30}, a number of similar approaches combining relational (logic-based) and deep learning methods arose under a similar name of ``deep relational learning''~\cite{vsir2021deep}. Nevertheless, for compatibility with the recently introduced frameworks~\cite{fey2024position}, we hereby continue with the contemporary RDL view, where RDBs are first transformed into a graph-based representation suitable for the GNN-based learning. 

\subsubsection{Database Representation} 
\label{sec:db-graph}

The fundamental characteristic of RDL~\cite{fey2024position} is to represent an RDB as a heterogeneous graph.\footnote{sometimes referred to as the ``relational entity graph''} 
The graph representation can be defined as $G = (\mathcal{V}, \mathcal{E}, \mathcal{T}^v, \mathcal{T}^e)$, where $\mathcal{V}$ is the set of nodes, $\mathcal{E}$ is the set of edges, $\mathcal{T}^v$ is a set of node types with a mapping $\phi: \mathcal{V} \to \mathcal{T}^v$, and  $\mathcal{T}^e$ is a set of edge types with a mapping $\psi: \mathcal{E} \to \mathcal{T}^e$. The node types and edge types collectively form the graph \textit{schema} $(\mathcal{T}^v, \mathcal{T}^e)$.

Given an RDB schema $\mathcal{R}$, the node types $T \in \mathcal{T}^v$ correspond to the relations (tables) $T$ within the database $\mathcal{T}^v \overset{1:1}{\to} \mathcal{R}$,
while the edge types $\mathcal{T}^e$ represent the undirected inter-relations between the tables, as defined by the primary-foreign key pairs:
$\mathcal{T}^e = \{({R_i, R_j})~|~{R_i}[FK_{R_j}] \subseteq {R_j}[PK]~\lor~{R_j}[FK_{R_i}] \subseteq {R_i}[PK]\} \text{.}$
For a specific \textit{instance} of an RDB $\mathcal{R}$, the set of nodes $\mathcal{V}$ is then defined as the union of all tuples (rows) $t_i$ from each relation
$\mathcal{V} = \{v_{i,j}~|~R_i \in \mathcal{R},~t_j \in R_i\} \text{,}$
and the set of edges $\mathcal{E}$ is defined as $\mathcal{E} = \{({v_{i,k}, v_{j,l}}) |~t_k \in R_i,~t_l \in R_j, (R_i, R_j) \in \mathcal{T}^e\}$.

The graph representation is further enriched by \textit{node embedding matrices}, \textit{attribute schema}, and optionally a \textit{time mapping}. {Node embedding matrix} $h^{(l)}_v \in \mathbb{R}^{d \times d_{\phi(v)}}$ contains the embedding representation of a node $v \in \mathcal{V}$ in a given layer $l$. With an {attribute schema} $\mathcal{A}_T$ that provides information about the types of attributes $A_1,\dots,A_n$ associated with the nodes $v$ of a specific node type $T \in \mathcal{T}^v$, the initial embedding tensors $h_v^{(0)} \in \mathbb{R}^{d^{(0)} \times n}$ are computed from the raw database attribute tuples ${t_i} = (a_1, a_2, \ldots, a_n)$ through multi-modal attribute encoders~\cite{fey2024position}. Finally, the \textit{time mapping} is a function $\tau$ that assigns a timestamp $t_v$ to each node $\tau: v \mapsto t_v $, effectively creating a dynamically growing graph in time, enabling the use of temporal graph sampling~\cite{rossi2020temporal}.

\subsubsection{Predictive Tasks}
In RDL, predictive tasks are implemented through the creation of dedicated training tables $T_t$ that extend the existing relational schema of $\mathcal{R}$. As introduced in~\cite{fey2024position}, a training table $T_t$ contains two essential components: foreign keys $T_t[FK]$ that identify the entities of interest and target labels $y \in \mathcal{A}_{T_t} \setminus T_t[FK]$. Additionally, timestamps $t_v \in \mathcal{A}_{T_t}$ that define temporal boundaries for the prediction of $y$ can also be included. 

The training table methodology supports a diverse range of predictive tasks, including node-level predictions (e.g., customer churn, product sales), link predictions between entities (e.g., user-product interactions), and, crucially, both temporal and static predictions. In the case of temporal predictions, a timestamp attribute $t_v$ in the training table $T_t$ specifies when the prediction is to be made, restricting the model to only consider information available up to the point $t_v$ in time.

\subsubsection{Neural Architecture Space} 
\label{sec:rdl-neural-space}
Building upon the heterogeneous graph representation $G$, RDL models generally consist of the following four major stages.

\begin{enumerate}
\item \textbf{Table-level attribute encoder} creates the initial node embedding matrices $h_v^{(0)} \in \mathbb{R}^{d^{(0)} \times n}$, i.e. sequences of $n$ embedding vectors $\mathbb{R}^{d^{(0)}_{\phi(v)}}$ for each attribute $A_1,\dots,A_n$ of $\phi(v)$ based on its respective semantic data type.

\item \textbf{Table-level tabular model} allows to employ existing tabular learning models~\cite{chen2023trompt,hu2020tabtransformer} to yield more sophisticated node embeddings $h_v^{(l)}$. Notably, in this stage, an RDL model \textit{may} reduce the dimensionality of the node attribute matrix embedding $h_v^{(l)} \in \mathbb{R}^{d^{(l)} \times n}$ to a vector embedding $h_v^{(l)} \in \mathbb{R}^{d^{(l)}_{\phi(v)}}$.


\item \textbf{Graph neural model} then depends on the chosen embedding dimensionality of $h_v^{(l)}$. If there is a single embedding vector $h_v^{(l)} \in \mathbb{R}^{d^{(l)}_{\phi(v)}}$ per each node, the model can employ standard GNN~(Sec.~\ref{sec:gnn}) heterogeneous message-passing~\cite{velivckovic2018graph,brody2022how}, otherwise a custom message-passing scheme~\cite{peleska_transformers_2024} is required.

\item \textbf{Task-specific model head} finally provides transformation of the resulting node embeddings into  prediction, usually involving simple MLP layers.
\end{enumerate}



\section{Database-Specific Task-Agnostic Pretraining}
Prior work in the emerging field of RDL has primarily focused on task-specific models trained using supervised learning~\cite{robinson2024relbench,peleska_transformers_2024,chen_relgnn_2025}. However, a single relational database often supports a wide range of predictive tasks, spanning node-level classification and regression, link prediction, and temporal forecasting. Adhering to the task-specific paradigm in such settings is both computationally inefficient and operationally burdensome, as each task requires a separate model trained from scratch with its own parameters.

To overcome this limitation, we propose a database-specific, task-agnostic pretraining approach based on self-supervised contrastive learning~\cite{baldi_contrastive_1991}. Our method learns transferable representations by capturing structural and semantic signals across the database graph, without reliance on downstream labels. Specifically, we introduce a three-level contrastive objective that operates on different granularities of the relational data: individual rows, inter-row links, and contextual neighborhoods. This enables the model to learn rich, reusable embeddings that generalize well across multiple tasks defined on the same database.

\subsection{Three-Level Contrastive Pretraining}
\label{sec:loss}
Contrastive pretraining has become a standard technique in both tabular learning~\cite{wang_transtab_2022,zhu_xtab_2023} and (heterogeneous) graph learning~\cite{jiang_pre-training_2021,hassani_contrastive_2020,li_augmentation-free_2024}. Since RDL inherently combines elements from both domains, it is natural to extend contrastive approaches to this setting as well.
To that end, we introduce a \textit{three-level contrastive pretraining framework}, designed to capture the heterogeneous structure of relational databases, where each level targets a different structural aspect:
\begin{enumerate}
    \item \textit{row-level}, capturing standard intra-tabular attribute (feature) patterns,
    \item \textit{link-level}, modeling direct inter-tabular referential relationships,
    \item and \textit{context-level}, representing broader relational context through wider neighborhoods in the derived database graph.
\end{enumerate}
This multi-level formulation then enables the model to learn general-purpose representations that are both semantically meaningful and structurally aware, supporting a wide range of downstream tasks.

\subsubsection{Row-Level Pretraining}
To learn robust embeddings of individual rows that reflect the heterogeneity across different database tables, we introduce a contrastive pretraining objective based on data corruption. Specifically, for each (sampled) raw database tuple $t_j = (a_1, a_2, \ldots, a_n)$ from a relation $R_i$, we generate a corrupted version $\hat{t}_j$ by randomly selecting a subset of attribute values with uniform probability $p$. The selected values are then replaced with values drawn from the empirical marginal distribution of the respective attributes, defined as a uniform distribution over the values observed in the training data. This approach is inspired by \textsc{Scarf}~\cite{bahri_scarf_2021}, but extended to the relational setting.

Importantly, primary and foreign key attributes are excluded from corruption to preserve the integrity constraints of the original database schema. We then follow the standard RDL pipeline (Sec.~\ref{sec:rdl}) by passing both the original and corrupted tuples, $t_j$ and $\hat{t}_j$, through the RDL model to obtain their respective node embeddings: $h_{v}$ and $\hat{h}_{v}$, where $v \sim t_j \in R_i$ denotes the corresponding node.

Given a node $v$ of type $\phi(v)$, we compute the similarity between the original embedding $h_{v}$ and its corrupted counterpart $\hat{h}_{v}$, and contrast it against the similarity to corrupted embeddings $\hat{h}_u$ of other nodes $u$ with the same type. This leads to the following InfoNCE-style~\cite{gutmann_noise-contrastive_2010} contrastive loss:

\begin{equation}
\label{eq:loss-row}
    \mathcal{L}^{\text{row}}_{v} = - \log{\frac{\exp{(\hat{h}_{v}^{\top} W_{\phi(v)} h_{v})}}{\sum\limits_{u \in \{v\} \cup \mathcal{V}^-_v} \exp{(\hat{h}_{u}^{\top} W_{\phi(v)} h_{v})}}} ,
\end{equation}

where $W_{\phi(v)} \in \mathbb{R}^{d \times d}$ is a \textit{learnable} similarity matrix specific to each node type, and $\mathcal{V}^-_v \subseteq \{w \mid \phi(w) = \phi(v),~w \neq v\}$ is a set of negative samples with cardinality $|\mathcal{V}^-_v| \leq N^-_{\text{max}}$.


\subsubsection{Link-Level Pretraining}
To capture the specific semantics of relationships between rows connected via primary and foreign key constraints, we introduce a contrastive learning objective at the level of individual edges. Each such relationship corresponds to a particular edge type in the graph representation of the RDB (Sec.~\ref{sec:rdl}).

Given a directed edge $e_{u,v}$ from source node $u$ to target node $v$ with edge type $\psi(e)$, we aim to distinguish the actual linked pair $(u, v)$ from randomly sampled negative pairs. Specifically, we compare the similarity between the true source and target embeddings $(h_u, h_v)$ to the similarity between $h_v$ and other nodes $h_w$ of the same type as $u$ that are not connected to $v$ via an edge of type $\psi(e)$. This leads to the following contrastive loss:

\begin{equation}
\label{eq:loss-link}
    \mathcal{L}^{\text{link}}_{e_{u,v}} = - \log{\frac{\exp{(h_{u}^{\top} W_{\psi(e)} h_{v})}}{\sum\limits_{w \in \{u\} \cup \mathcal{V}^-_{e_{u,v}}} \exp{(h_{w}^{\top} W_{\psi(e)} h_{v})}}} ,
\end{equation}

where $W_{\psi(e)} \in \mathbb{R}^{d \times d}$ is a \textit{learnable} similarity matrix specific to each edge type $\psi(e)$, and $\mathcal{V}^-_{e_{u,v}} \subseteq \{w \mid \phi(w) = \phi(u),~w \neq u,~(w, v) \notin \mathcal{E}_{\psi(e)}\}$ is a set of negative samples of cardinality $|\mathcal{V}^-_{e_{u,v}}| \leq N^-_{\text{max}}$.



\subsubsection{Context-Level Pretraining}
While the link-level loss captures local relational dependencies between pairs of nodes, it remains limited to direct connections. To model more complex, higher-order structural interactions, we introduce a \textit{context-level} contrastive objective.

For each node $v$ of type $\phi(v)$, we first compute its \textit{context embedding} $c_v$ as the average of transformed embeddings of its neighboring source nodes:
\begin{equation}
\label{eq:context}
   c_{v} = \frac{1}{|\mathcal{N}_v|} \sum_{u \in \mathcal{N}_v} h_u^{\top} W_{\phi(u)} ,
\end{equation}
where $\mathcal{N}_v$ is the set of all source neighbors of node $v$, and $W_{\phi(u)} \in \mathbb{R}^{d \times d}$ is a \textit{learnable} transformation matrix specific to the node type $\phi(u)$. Notably, the transformation matrices used here are distinct from those used in the row-level loss (Eq.~\ref{eq:loss-row}), allowing the model to specialize its representations for context aggregation.

Next, we compute the similarity between the input node embedding $h_v$ and its aggregated context embedding $c_v$, and contrast it against similarities with context embeddings of other nodes of the same type. This yields the following context-level contrastive loss:
\begin{equation}
\label{eq:loss-context}
    \mathcal{L}^{\text{context}}_{v}  = - \log{\frac{\exp{(c_{v}^{\top} h_{v})}}{\sum\limits_{u \in \{v\} \cup \mathcal{V}^-_v} \exp{(c_{u}^{\top} h_{v})}}} ,
\end{equation}
where $\mathcal{V}^-_v \subseteq \{w \mid \phi(w) = \phi(v),~w \neq v\}$ is a set of negative samples with cardinality $|\mathcal{V}^-_v| \leq N^-_{\text{max}}$.

\subsection{Pretraining Pipeline}

\subsubsection{Sampling}
The sheer scale of many relational databases, which often contain millions of interconnected rows~\cite{motl2015ctu,robinson2024relbench}, makes training on their full graph representations computationally infeasible. This necessitates sampling smaller, tractable subgraphs for training. While neighborhood sampling~\cite{hamilton2017inductive} based on breadth-first search is a common technique for tasks centered on a single node type predictions (e.g.,  customer churn), it is ill-suited for our pretraining objectives.
On heterogeneous graphs, such as those generated from RDBs, neighborhood sampling creates subgraphs with a heavily skewed distribution of node and edge types. This imbalance is problematic for our global, task-agnostic objectives, which rely on a balanced view of the graph's structure (Sec.~\ref{sec:loss}). To address this, we employ a variant of HGSampling~\cite{Hu2020}, specifically designed to create subgraphs with an equal representation of node and edge types.

\subsubsection{Combined Loss}
The proposed contrastive losses are designed to encapsulate the heterogeneous nature of relational databases; however, this very heterogeneity makes it difficult to combine them into a single loss function for training the RDL model. The challenge primarily arises from the varying number of entities (rows) involved in the computation of each loss component. While HGSampling can partially address these discrepancies, a fundamental difference in scale remains between the loss components: the link-level loss is driven by edges (links), whereas the row-level and context-level losses are driven by nodes (rows).

To mitigate the variance between these components, we apply a dynamic normalization factor, $\mu$, based on the number of negative samples utilized for a given input. This factor is defined as:

\begin{equation}
\label{eq:norm-factor}
    \mu(N^{-}) = -\log{\frac{1}{N^{-} + 1}} ,
\end{equation}
where $N^{-}$ is the number of negative samples used in the loss computation.

For a graph representation $G = (\mathcal{V}, \mathcal{E}, \mathcal{T}^v, \mathcal{T}^e)$ of a relational database and a sampled subgraph $G' = (\mathcal{V'}, \mathcal{E'}, \mathcal{T}^v, \mathcal{T}^e)$, we define the combined loss for the subgraph as:

\begin{equation}
\label{eq:full-loss}
    \mathcal{L} = \frac{1}{|\mathcal{V'}|} \sum_{v \in \mathcal{V'}}{\frac{\mathcal{L}^{\text{row}}_v}{\mu_v}} + \frac{1}{|\mathcal{E'}|} \sum_{e_{u,v} \in \mathcal{E'}}{\frac{\mathcal{L}^{\text{link}}_{e_{u,v}}}{\mu_{e_{u,v}}}} + \frac{1}{|\mathcal{V'}|} \sum_{v \in \mathcal{V'}}{\frac{\mathcal{L}^{\text{context}}_v}{\mu_v}} ,
\end{equation}
where $\mu_v = \mu(|\mathcal{V}^-_{v}|)$ and $\mu_{e_{u,v}} = \mu(|\mathcal{V}^-_{e_{u,v}}|)$.

\section{Experiments}
\label{sec:experiments}

In this section we present our initial experimental assessment of the proposed pretraining methodology and discuss its strengths and weaknesses.

\subsection{Experimental Setup}
We evaluate our method over a range of hyperparameters affecting the model architecture and the learning process itself. As the overall approach is significantly resource-intensive, we focus our evaluation on two databases from the established \textsc{RelBench} benchmark~\cite{robinson2024relbench}, whose characteristics are detailed in Table~\ref{tab:selected_dbs}.

\begin{table}[t]
    \centering

\begin{tabularx}{\textwidth}{L{5}C{2}C{2}C{2.5}C{3}C{3.4}C{2.5}C{3}C{2.4}C{2.4}C{2.5}C{2.5}}
    \hline
    \toprule
    
    \textbf{Dataset} & \textbf{Tab} & \textbf{FK} & \textbf{Cols} & \textbf{Rows} & \textbf{Links} & \textbf{Loop} & \textbf{Dom.} & \textbf{Cat} & \textbf{Num} & \textbf{Text} & \textbf{Time} \\
    \midrule
    rel-f1 & 9 & 13 & 45 & 97.6k & 227.7k & \cmark & Sport & 16 & 13 & 9 & 7 \\
    rel-stack & 7 & 12 & 33 & 5.4M & 7.5M & \cmark & Edu. & 10 & 1 & 13 & 7 \\
    \hline
\end{tabularx}
    \caption{Characteristic features of the databases used for experiments.}
    \label{tab:selected_dbs}
\end{table}

\subsubsection{Backbone RDL Models}
The backbone models used for the experiments follow the RDL blueprint and the outlined neural architecture space, as described in Sec.~\ref{sec:rdl-neural-space}. Nevertheless, the backbone models incorporate only the first three stages, thereby omitting the task-specific model head. To ensure comparability across the experiments, we present two models featuring distinct table-level architectures while utilizing the same attribute encoders for numerical, categorical, multi-categorical, textual, and temporal values, as well as the same graph neural model.

\paragraph{GraphSAGE with Linear Transformation} 
This model applies a linear transformation on top of a concatenation of the attribute $a_1,\dots,a_n$ embeddings $h_v^{(0)} \in \mathbb{R}^{n \cdot d^{(0)}}$ to yield a single embedding vector $h_v^{(1)} = {W} h_v^{(0)}$ for each node $v$. The projected node embeddings $h^{(1)} \in \mathbb{R}^{d_{\phi(v)}}$ then form input into the GraphSAGE~\cite{hamilton2017inductive} model, forming the GNN level.

\paragraph{GraphSAGE with Tabular ResNet}
This model is similar to the previous, but it employs a more sophisticated tabular ResNet model~\cite{gorishniy_revisiting_2021} for the tabular-level stage to reduce the dimensionality of the node embeddings before passing them to the GraphSAGE layer.


\subsubsection{Pretraining}


In our pretraining experiments, we trained the backbone models by optimizing the proposed three-level contrastive loss (Eq.~\ref{eq:full-loss}) using the Adam optimizer~\cite{Kingma2015adam} with a learning rate of $0.001$. To prevent data leakage from the validation and test sets, we pruned the database to contain only data available up to the predefined validation timestamp from \textsc{RelBench}.

We sampled the database's graph representation using HGSampling, seeded with a single input node type corresponding to the table with the highest number of foreign keys (i.e., `results` for `rel-f1` and `posts` for `rel-stack`). HGSampling was configured to sample 64 nodes for each node type over 3 iterations. Internally, we set the maximum number of negative samples per input $N_{max}^-$ to a constant value of 256 for all loss levels. All models had 128 internal hidden channels and were evaluated over a hyperparameter grid, that included the number of message-passing layers (2 or 3), the aggregation function (summation or mean), and the cell value corruption probability (0.2, 0.4, or 0.6).

The models were trained with a batch size of 64  for a maximum of 2000 steps, with a time limit of 4 hours.
We employed an early stopping procedure with validation performed every 50 steps and a patience of 10 rounds. Each validation was performed on 50 new samples, randomly sampled from the database using the described HGSampling setup.

\subsubsection{Task-Specific Fine-Tuning}
The pretrained models were further trained on the standardized supervised downstream tasks from \textsc{RelBench}~\cite{robinson2024relbench}, which include entity binary classification and regression. For each task, we equipped the backbone model with a distinct, task-specific head formed by an MLP with a single hidden layer of 128 channels, batch normalization and a ReLU activation function. The model was then trained by optimizing either the Cross-Entropy loss for binary classification or the Mean Squared Error loss for regression, respectively, using the Adam optimizer with a learning rate of $0.0001$.

Similar to the pretraining phase, we limited the training to a maximum of 2000 steps and a time limit of 2 hours.
We also employed an early stopping procedure, with validation performed every 100 steps and a patience of 5 epochs. For graph sampling, we used Neighborhood Sampling with a batch size of 512 and 128 neighbors per node, with a sampling depth equal to the number of the model's message-passing layers.




\subsection{Results}

\begin{table}[t]
    \centering
    \renewcommand{\arraystretch}{1.1}
\begin{tabularx}{\textwidth}{L{5}L{8}C{3.2}|M{1cm}M{1cm}M{1cm}M{1cm}M{1cm}M{1cm}}
\hline
\toprule
\multicolumn{9}{c}{\textbf{Binary Classification}} \\
\multicolumn{9}{c}{Model AUC-ROC} \\

\hline

\multirow[c]{2}{*}{\textbf{Dataset}} & \multirow[c]{2}{*}{\textbf{Task}}  & \multirow[c]{2}{*}{\textbf{Split}}  & \multicolumn{3}{c}{\textbf{Linear SAGE}} & \multicolumn{3}{c}{\textbf{ResNet SAGE}} \\ 
 &  &  & B & P & P\&F & B & P & P\&F \\
\hline
\multirow[c]{4}{*}{rel-f1} & \multirow[c]{2}{*}{driver-dnf} & Val & 72.91 & 74.39 & 73.25 & 74.69 & 74.59 & 78.35 \\
 &  & Test & 75.58 & 72.93 & 73.76 & 78.07 & 76.57 & 78.69 \\
\cline{2-9}
 & \multirow[c]{2}{*}{driver-top3} & Val & 78.02 & 79.67 & 78.91 & 77.40 & 82.51 & 82.16 \\
 &  & Test & 84.60 & 79.16 & 82.24 & 84.57 & 82.11 & 71.37 \\
\hline
\multirow[c]{4}{*}{rel-stack} & \multirow[c]{2}{*}{user-badge} & Val & 89.35 & 87.72 & 89.63 & 89.29 & 87.25 & 89.46 \\
 &  & Test & 88.12 & 85.96 & 88.51 & 87.89 & 85.56 & 88.21 \\
\cline{2-9}
 & \multirow[c]{2}{*}{user-engagement} & Val & 90.02 & 86.93 & 90.08 & 89.94 & 87.06 & 90.18 \\
 &  & Test & 90.34 & 86.99 & 90.47 & 90.28 & 86.55 & 90.32 \\ 
\hline
\toprule
\multicolumn{9}{c}{\textbf{Regression}} \\
\multicolumn{9}{c}{Model MAE} \\

\hline
\multirow[c]{2}{*}{rel-f1} & \multirow[c]{2}{*}{driver-position} & Val & 3.202 & 3.227 & 3.118 & 3.113 & 3.208 & 3.076 \\
 &  & Test & 3.686 & 3.424 & 3.591 & 3.329 & 3.581 & 3.394 \\
\hline
\multirow[c]{2}{*}{rel-stack} & \multirow[c]{2}{*}{post-votes} & Val & 0.099 & 0.104 & 0.079 & 0.089 & 0.102 & 0.085 \\ 
 &  & Test & 0.105 & 0.110 & 0.086 & 0.095 & 0.108 & 0.090 \\
\hline
\end{tabularx}

    \caption{Best results on the entity binary classification (AUC-ROC, higher is better) and regression tasks (MAE, lower is better).}
    \label{tab:results}
\end{table}

Table~\ref{tab:results} presents the main results of our experiments, using the Area Under the Receiver Operating Characteristic Curve (AUC-ROC) metric for the binary classification tasks and the Mean Absolute Error (MAE) metric for the regression tasks, respectively. We evaluated the models in three distinct scenarios to assess the impact of our pretraining method:
\begin{enumerate}
    \item A \textit{baseline} model (B) trained from scratch on the downstream task.
    \item A \textit{pretrained} model (P) where the backbone parameters were \textit{frozen}, and only the task-specific head was trained.
    \item A \textit{pretrained} model that was fully \textit{fine-tuned} (P\&F) on the downstream task, updating the parameters of both the backbone and the head.
\end{enumerate}

Our results show that on the smaller `rel-f1` dataset, the frozen pretrained model (P) provides promising performance, performing comparably to the baseline model (B) and even outperforming it on the `driver-top3` task within a validation split. Nevertheless, on the larger `rel-stack` dataset, the same frozen models fell behind the performance of the baseline. This might be due to a number of reasons, such as the limited capacity of the model, too short a training period, or a missed aspect of the data, which we discuss further in Sec.~\ref{sec:future-work}.

The pretrained and fine-tuned models (P\&F), however, consistently outperformed the baseline models (B) on the validation splits across all tasks. Importantly, these models provided a significant performance boost on the larger `rel-stack` database, highlighting the ability of our pretraining procedure to learn meaningful and transferable representations, agnostic of the downstream task.

\section{Conclusion}
\label{sec:conclusion}
In this work, we introduced a self-supervised pretraining method for representation learning on relational databases, building on the emerging paradigm of Relational Deep Learning (RDL). The approach learns effective relational representations by implementing a three-level contrastive objective---at the row, link, and context levels---without the need for downstream supervision. The evaluated RDL models, coupled with type-balanced heterogeneous graph sampling, demonstrated promising results across both classification and regression tasks. Particularly, our findings indicate that pretraining offers measurable benefits, especially in low-data regimes, and generally enhances model performance when fine-tuned, paving the way towards the development of foundational models for relational data.

\subsubsection{Future Work}
\label{sec:future-work}
Despite these promising initial findings, the scope of our experiments was significantly restricted, and a more thorough study is needed to fully assess the capabilities of the pretraining regime. Notably, pretrained models without the finetuning on the downstream task fell behind in performance on the larger dataset. We hypothesize that this limitation is due to capacity constraints of the model and, in the future, we would like to enhance the experiments with larger Transformer-based models with higher representation dimensionality~\cite{peleska_transformers_2024}. Additionally, our pretraining method currently neglects the temporal dimension of relational data, and incorporating time-aware mechanisms is a promising direction for encoding more complex relationships. Finally, extending the pretraining paradigm to support cross-database generalization or continual learning could dramatically improve its usability in real-world applications.

\bibliographystyle{splncs04}
\bibliography{bibliography}

\end{document}